\documentclass{llncs}
\usepackage{makeidx}  
\usepackage{url}      
\usepackage{amsmath}
\usepackage{url}

\usepackage{graphicx}
\graphicspath{{pics/}{figs/}}
\def\B#1{\mathbf{#1}}

\begin{document}
\frontmatter          
\pagestyle{empty}  
\mainmatter              
\title{CanICA: Model-based extraction of reproducible group-level ICA 
patterns from fMRI time series}
%
%
%
%
%

\author{G. Varoquaux$^1$, S. Sadaghiani$^2$, J.B. Poline$^2$, B. Thirion$^1$}
\authorrunning{Ga\"el Varoquaux} 
\tocauthor{Ga\"el Varoquaux}
\institute{INRIA, Saclay-Ile de France, Saclay, France, \and
CEA/Neurospin, Saclay, France
\thanks{Funding from INRIA-INSERM collaboration. 
The fMRI set was acquired in the
context of the SPONTACT ANR project}
}

\maketitle 
\vspace*{-1em}

\begin{abstract}

Spatial Independent Component Analysis (ICA) is an increasingly used
data-driven method to analyze functional Magnetic Resonance Imaging
(fMRI) data. To date, it has been used to extract meaningful patterns
without prior information. However, ICA is not robust to mild data
variation and remains a parameter-sensitive algorithm. The validity of
the extracted patterns is hard to establish, as well as the significance
of differences between patterns extracted from different groups of
subjects. We start from a generative model of the fMRI group data to
introduce a probabilistic ICA pattern-extraction algorithm, called CanICA
(Canonical ICA). Thanks to an explicit noise model and canonical
correlation analysis, our method is auto-calibrated and identifies the
group-reproducible data subspace before performing ICA. We compare our
method to state-of-the-art multi-subject fMRI ICA methods and show that
the features extracted are more reproducible.

\end{abstract}

\section{Introduction}
\label{sec:intro}

Resting-state fMRI is expected to give insight into the intrinsic
structure of the brain and its networks. In addition, such protocols can
be easily applied to impaired subjects and can thus yield useful
biomarkers to understand the mechanisms of brain diseases and for
diagnosis. Spatial ICA has been the most successful method for
identifying meaningful patterns in resting-state fMRI data without prior
knowledge. The use of the resulting patterns is widespread in cognitive
neuroscience, as they are usually well-contrasted, separate different
underlying physiological, physical, and cognitive processes, and bring
into light relevant long-range cognitive networks.

However, validation of the resulting individual patterns suffers from the
lack of testable hypothesis. As a result, cognitive studies seldom rely
on automatic analysis, and relevant ICA maps are cherry picked by eye to
separate them from noise-induced patterns. Probabilistic ICA models have
been used to provide pattern-level noise-rejection criteria
\cite{Beckmann2004} or likelihood for the model \cite{Guo2008}, but have
yet to provide adequate auto-calibration and pattern-significance
testing. The lack of reproducibility is detrimental to group analysis:
various, often non-overlapping, patterns have been published
\cite{Damoiseaux2006,Perlbarg2008a} and the statistical frameworks for
comparison or inference on ICA maps have to be further developed.

To allow for group analysis, it is important to extract from a group of
subjects ICA maps that are well-represented in the group. Various
strategies have been adopted: group ICA \cite{Calhoun2001a} concatenates
individual time series; tensor ICA \cite{Beckmann2005a} estimates ICA
maps across subjects with different loadings per subject; NEDICA
\cite{Perlbarg2008a} merges ICA maps by hierarchical clustering. 

In this paper, we present a novel model and a method, that we dub CanICA,
to extract only the reproducible ICA maps from group data. The strength
of this method lies in the elimination of the components non-reproducible
across subjects using model-informed statistical testing and canonical
correlation analysis (CCA). We compare the reproducibility of features
extracted by our method to features extracted using tensor ICA and group
ICA, but could not compare to NEDICA since it suppresses in some way
between-subject variability.


\section{Methods}

\label{sec:method}

\subsection{Generative model: from group-level patterns to observations}

At the group level, we describe intrinsic brain activity by a set of
spatial patterns $\B{B}$ corresponding to networks common to the group.
We give a generative model to account for inter-subject variability and
observation noise.

The activity recorded on each subject $s$ can be described by a
set of subject-specific spatial patterns $\B{P}_s$, which are a
combination of the group-level patterns $\B{B}$ with loadings given by
$\B{\Lambda}_s$ and additional subject-variability denoted as a residual matrix
$\B{R}_s$. If we write the spatial patterns $\B{B}$, $\B{P}_s$ and
$\B{R}_s$ as $n_\text{patterns} \times n_\text{voxels}$ matrices, for
each subject $s$, $ \B{P}_s = \B{\Lambda}_s \, \B{B} + \B{R}_s$. In other
words, at the group level, considering the group of patterns (vertically
concatenated matrices) $\B{P} = \{\B{P}_s\}$, $\B{R} = \{\B{R}_s\}$,
and $\B{\Lambda} = \{\B{\Lambda}_s\}$, $s = 1 \dots S$,
\begin{equation}
    \B{P} = \B{\Lambda} \, \B{B} + \B{R}.
\label{eq:subject_variability}
\end{equation}
For paradigm-free acquisitions, there is no specific time course set by
an external stimulus or task, and for each acquisition-frame time point a
mixture of different processes, described by different patterns, is
observed. The observed fMRI data is a mixture of these patterns
confounded by observation noise: let $\B{Y}_s$ be the resulting spatial
images in BOLD MRI sequences for subject $s$ (an $n_\text{frames} \times
n_\text{voxels}$ matrix), $\B{E}_s$ the observation noise, and $\B{W}_s$
a loading matrix such that:
\begin{equation}
    \B{Y}_s =  \B{W}_s \, \B{P}_s + \B{E}_s.
\label{eq:observation_noise}
\end{equation}

%

\subsection{Estimating group-reproducible patterns from fMRI data} 

\subsubsection{Noise rejection using the generative model.}

Starting from fMRI image sequences, $\{\B{Y}_s, s=1...S\}$, time-sliced
interpolated and registered to the MNI152 template, we separate
reproducible patterns from noise by estimating successively each step of
the above hierarchical model.

First, we separate observation noise $\B{E}_s$ from subject-specific
patterns $\B{P}_s$ (Eq. \ref{eq:observation_noise}) through principal
component analysis (PCA). The principal components explaining most of the
variance for a given subject's data set form the patterns of interest,
while the tail of the spectrum is considered as observation noise. Using
a singular value decomposition (SVD), $\B{Y}_s = \B{U}_s \, \B{\Sigma}_s
\, \B{V}^T_s$. The $n$ first columns of $\B{V}^T_s$ constitute the
``whitened'' patterns $\B{\hat{P}}_s$ that we retain:  $\B{\hat{P}}_s =
(\B{V}_s)_{1\dots n}$, and the residual constitutes the observation
noise: $\B{\hat{E}}_s = \B{Y}_s - (\B{U}_s \, \B{\Sigma}_s \,
\B{V}^T_s)_{1\dots n}$.

As PCA can be interpreted as latent-variable separation under the
assumption of normally-distributed individual random variates, we model
observation noise as normally-distributed. Following \cite{Mei2008}, we
set the number $n$ of significant PCA components retained by drawing a
sample null-hypothesis dataset using a random normal matrix and comparing
the bootstrap stability of PCA patterns for the measured signal and the
null-hypothesis sample. Unlike information-based criteria used in
previous methods \cite{Beckmann2004,Calhoun2001a} for order selection,
such as approximations of model evidence or BIC, the selected number of
significant components does not increase when adding artificially number
of noise sources\cite{Mei2008}, and thus does not diverge with long fMRI
time series. This is important to avoid extracting group-reproducible
patterns from observation noise.


To identify a stable-component subspace across subjects, estimating Eq.
(\ref{eq:subject_variability}), we use a generalization of canonical
correlation analysis (CCA). CCA is used to identify common subspaces
between two different datasets. While there is no unique generalization
to multiple datasets, an SVD of the various whitened and
concatenated datasets can be used and is equivalent to standard CCA in the
two-datasets case \cite{Kettenring1971}. Given $\B{\hat{P}} =
\{\B{\hat{P}}_s\}$, SVD yields $\B{\hat{P}} = \B{\Upsilon} \, \B{Z}
\, \B{\Theta}^T$ where $\B{\Theta}^T$ forms the canonical variables, and
$\B{Z}$ the canonical correlations, which yield a measure of
between-subject reproducibility. Estimation of the inter-subject
reproducible components $\B{\hat{B}}$ is given by the vectors of
$\B{\Theta}^T$ for which the corresponding canonical correlation
$\B{Z}$ is above significance threshold. $\B{\hat{\Lambda}}$ is
identified as the corresponding loading vectors of
$\B{\Upsilon}\,\B{Z}$. For a given number of selected components,
this estimator minimizes the sum of squares for the residual $\B{\hat{R}}
= \B{\hat{P}} - \B{\hat{\Lambda}} \, \B{\hat{B}}$.

The significance threshold on the canonical correlations is set by
sampling a bootstrap distribution of the maximum canonical correlation
using $\B{\hat{E}}_s$, the subject observation noise identified
previously, instead of $\B{\hat{P}}_s$. Selected canonical variables have
a probability $p < 0.05$ of being generated by the noise.

\subsubsection{Identifying independent features in the
group-level patterns}

The selected group-level components $\B{B}$ form reproducible patterns of
task-free activation, but they represent a mixture of various processes
and are difficult to interpret for lack of distinguishable shape standing
out. We perform source separation using spatial ICA on this subspace.
From the patterns $\B{B}$ we estimate a mixing matrix $\B{M}$ and
group-level independent components $\B{A}$ using the FASTICA algorithm
\cite{Hyvarinen2000}: $\B{B} = \B{M}\,\B{A}$. FASTICA is an optimization
algorithm which successively extracts patterns with maximally
non-Gaussian marginal distributions. This separation corresponds to
identifying the maximum-contrast patterns in the subspace of interest.
These spatial patterns correspond to minimally-dependant processes and
contain identifiable physiological, physical, or neuronal components. 



Consistent with the FASTICA model, the main regions that form the nodes
of the functional networks within the resulting patterns are
the regions with values corresponding to the non-Gaussian tails of the
histogram. Following \cite{Schwartzman2009}, we model the non-interesting
voxels as normally-distributed and estimate the null distribution from
the central part of the histogram. The voxels of interest are selected
using an uncorrected p-value of $10^{-3}$. We use a specificity criterion
rather than a false discovery rate as it yields more stable results,
especially on the very-long-tailed distributions that we encounter. As
the total number of voxels in the brain is about $40\,000$, we expect no
more than 40 false positives, which corresponds to a small
false-discovery rate. 

\subsection{Model validation for inter-subject generalization}

The validation criteria for an ICA decomposition are unclear, as this
algorithm is not based on a testable hypothesis. The use of ICA is
motivated by the fact that the patterns extracted from the fMRI data
display meaningful features in relation to our knowledge of functional
neuroanatomy. These features should be comparable between subjects, and
thus generalize to new subjects.


To test the reproducibility of the results across subjects, we split our
group of subjects in two and learn ICA maps from each sub-group: this
yields $\B{A}_1$ and $\B{A}_2$. We compare the overlap of thresholded
maps and reorder one set to match maps by maximum overlap.
Reproducibility can be quantified by studying the cross-correlation
matrix $\B{C} = \B{A}^T_1 \B{A}_2$. For unit-normed components,
$\B{C}_{i,j}$ is 1 if and only if $(\B{A}_1)_i$ and $(\B{A}_2)_j$ are
identical.

We define two measures for overall stability and reproducibility of the
maps. First, a measure of the overlap of the subspaces selected in both
groups is given by the energy of the matrix: $E = \text{tr}\,({\B{C}^T
\B{C}})$. To compare this quantity for different subspace sizes, we
normalize it by the minimum dimension of the subspaces, $d =
\min(\text{rank}\,\B{A}_1, \text{rank}\,\B{A}_2)$: $e = \frac{1}{d}
\text{tr}\,({\B{C}^T \B{C}})$. $e$ quantifies the reproducibility of the
subspace spanned by the maps. For $e=1$, the two groups of  maps span the
same subspace, although individual independent components may differ.
Second, we use an overall measure of reproducibility for the maps: the
normalized trace of the reordered cross-correlation matrix $\B{C}$, $t =
\frac{1}{d}\text{tr}\,(\B{C})$. Indeed, after $\B{A}_2$ has been
reordered to maximize matching with $\B{A}_1$, the diagonal coefficients
of $\B{C}$ give the overlap between matched components.
Finally, the maximum value of each row and column of $\B{C}$ expresses
the best match each component learned in one group on the set learned in
the other. We plot its histogram. This indicator accounts for components
of one group matching multiple components of the other.

\section{Results}
\label{sec:results}

\subsection{Extracted group-level patterns of interest}

\begin{figure}[t]
  \hspace*{-.05\linewidth}
  \begin{minipage}{1.1\linewidth}
  \begin{minipage}{.49\linewidth}
    \begin{tabular}{p{.7\linewidth}p{.3\linewidth}}
    \includegraphics[width=\linewidth]{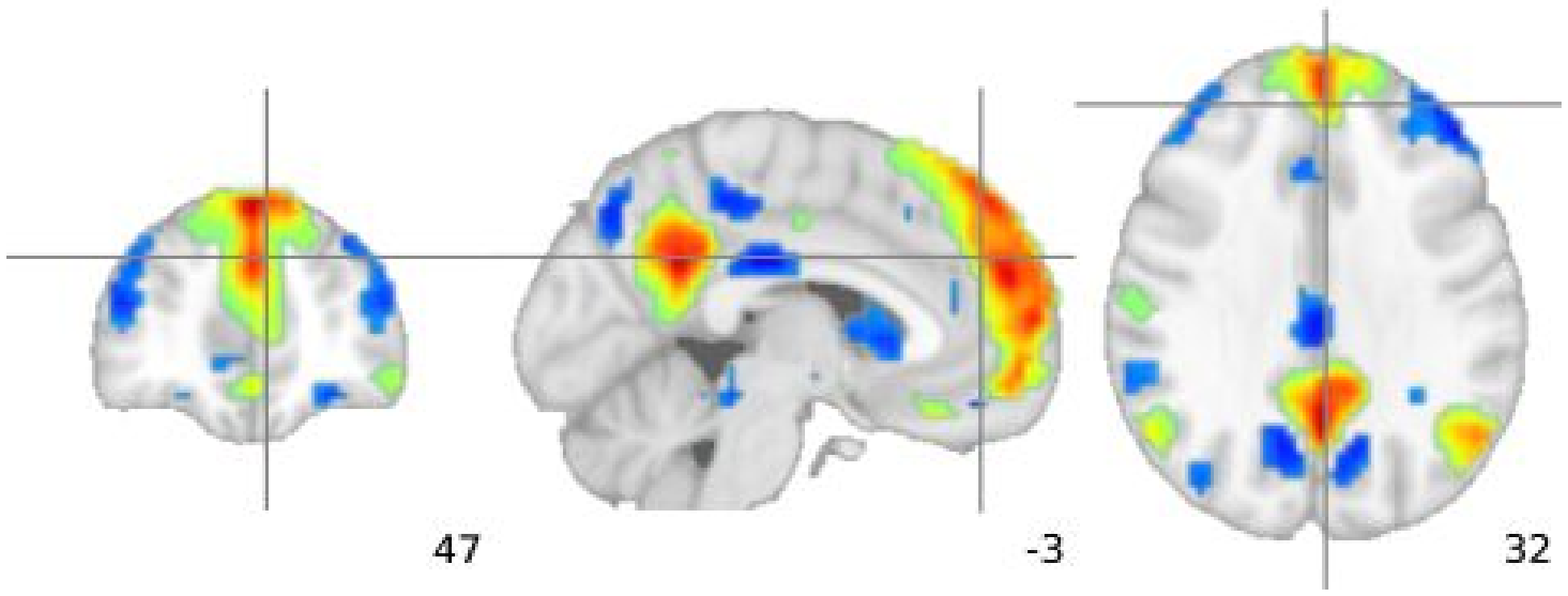} 
    \vskip-1.4em
    {~~\footnotesize\sf\bf (a)}
    &
    \hspace*{-.5ex}\includegraphics[width=\linewidth]{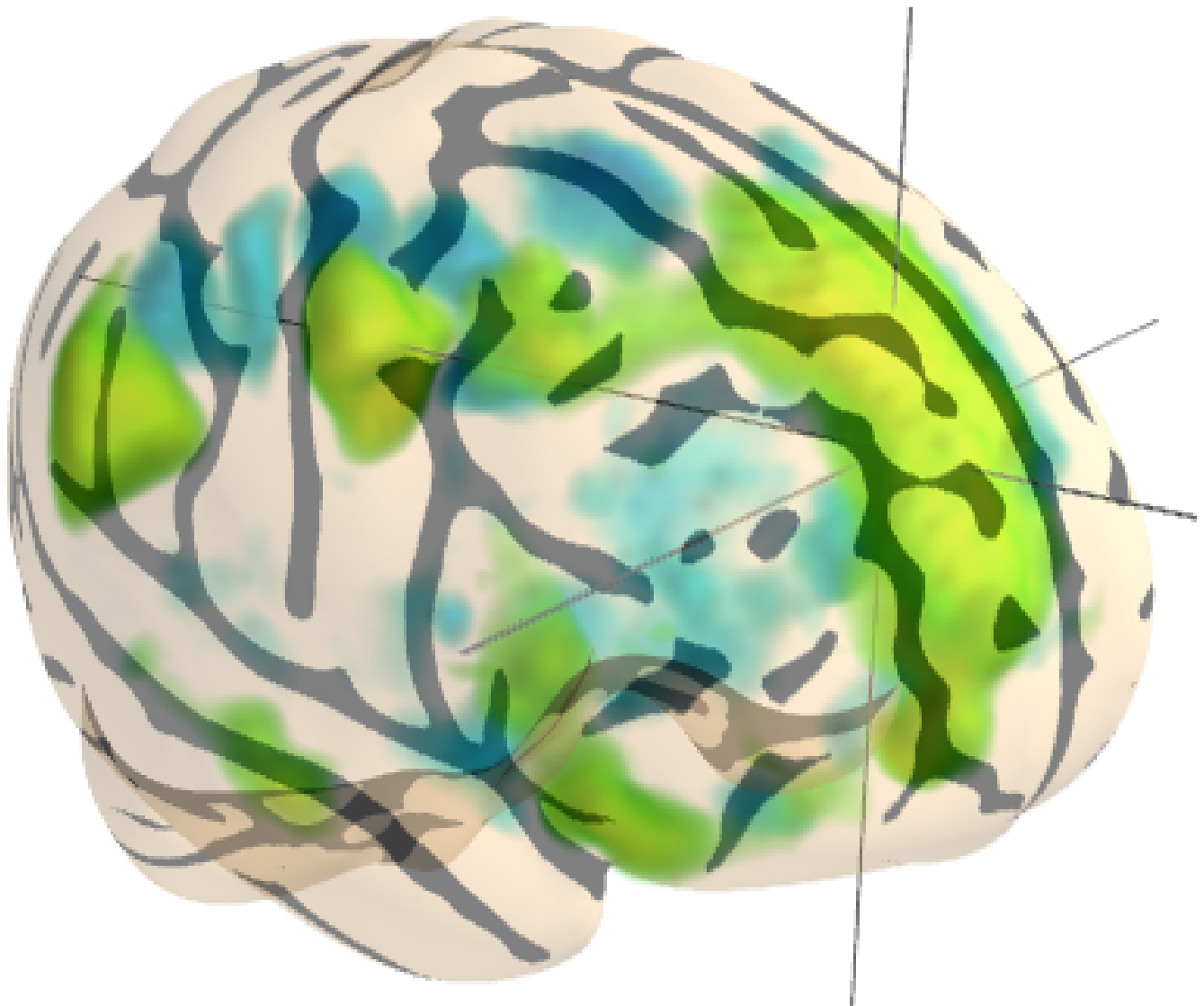} 
    \end{tabular}
  \end{minipage}%
  \hfill%
  \begin{minipage}{.49\linewidth}
    \begin{tabular}{p{.7\linewidth}p{.3\linewidth}}
    \includegraphics[width=\linewidth]{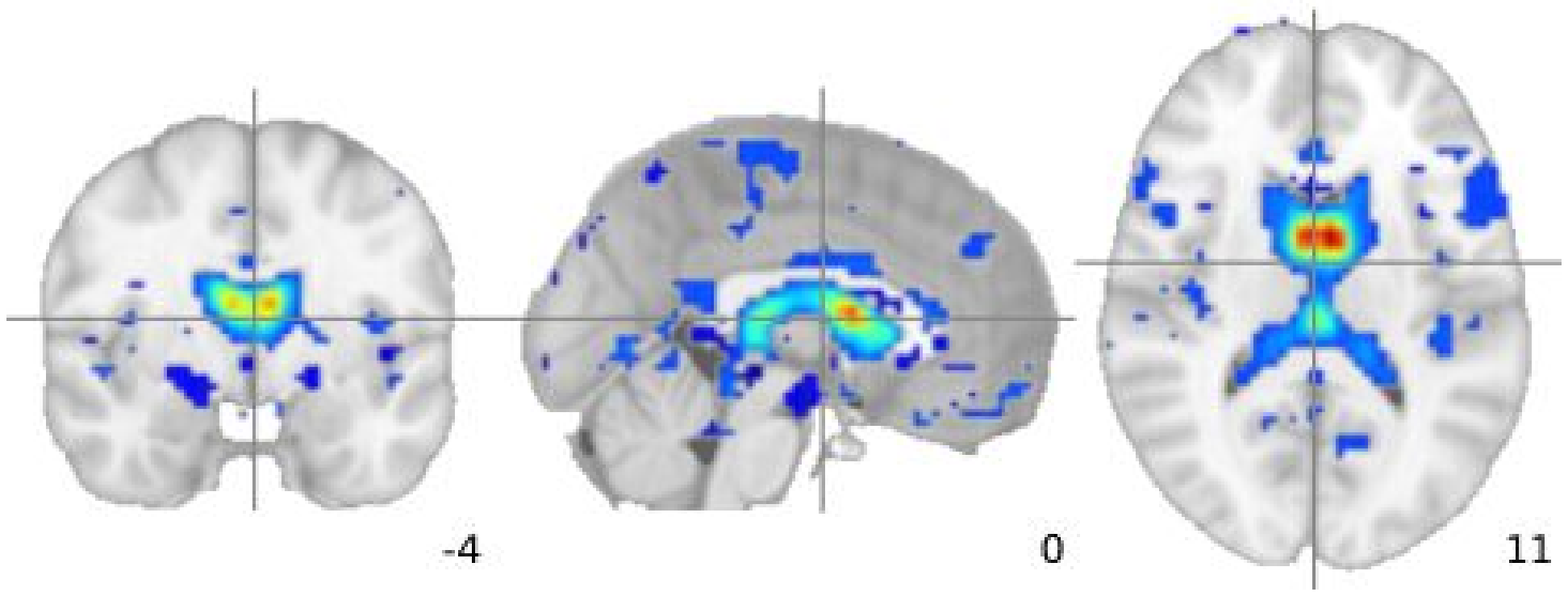} 
    \vskip-1.4em
    {~\footnotesize\sf\bf (b)}
    &
    \hspace*{-.5ex}\includegraphics[width=\linewidth]{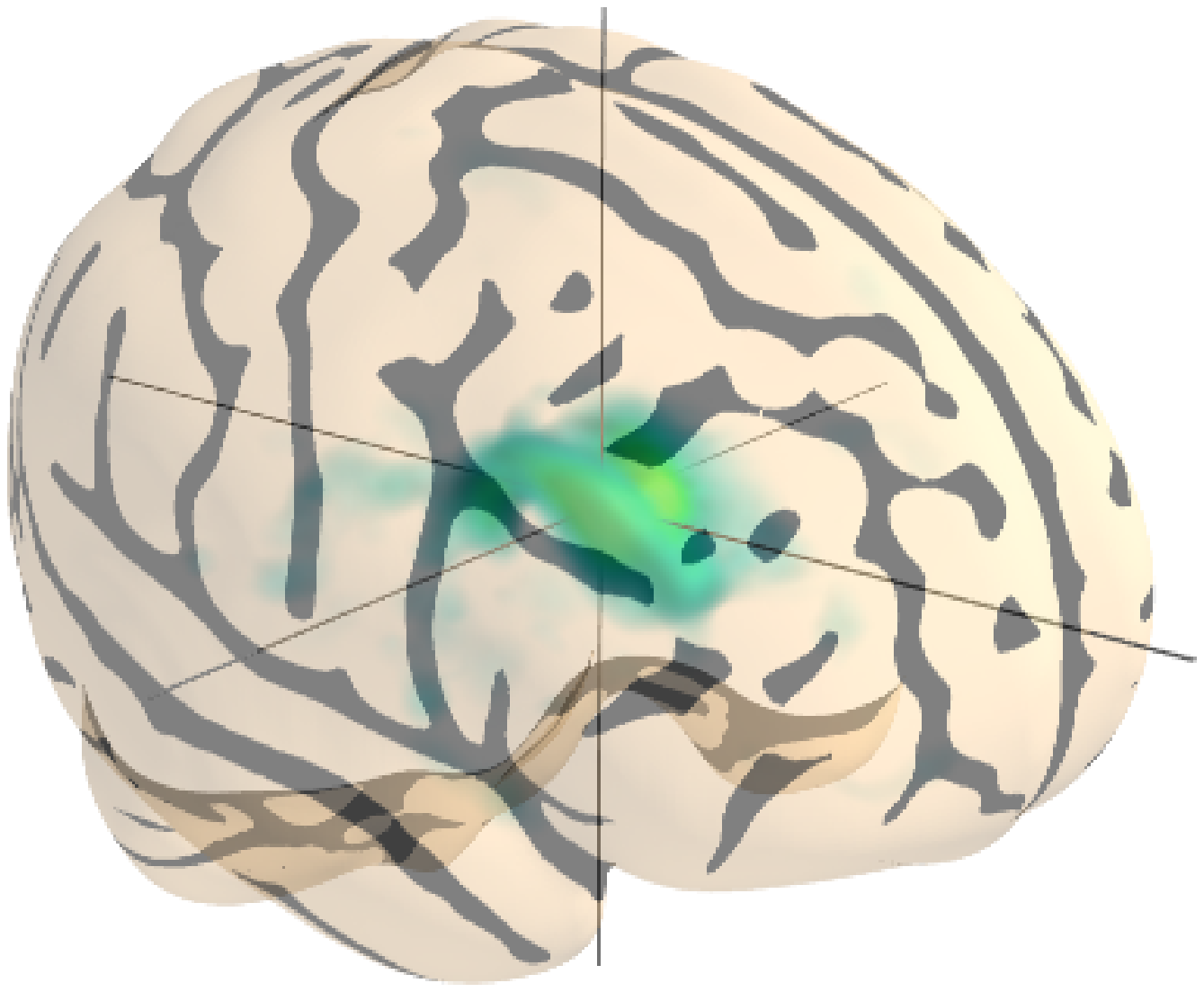} 
    \end{tabular}
  \end{minipage}%
  \vskip-1ex
  \begin{minipage}{.49\linewidth}
    \begin{tabular}{p{.7\linewidth}p{.3\linewidth}}
    \includegraphics[width=\linewidth]{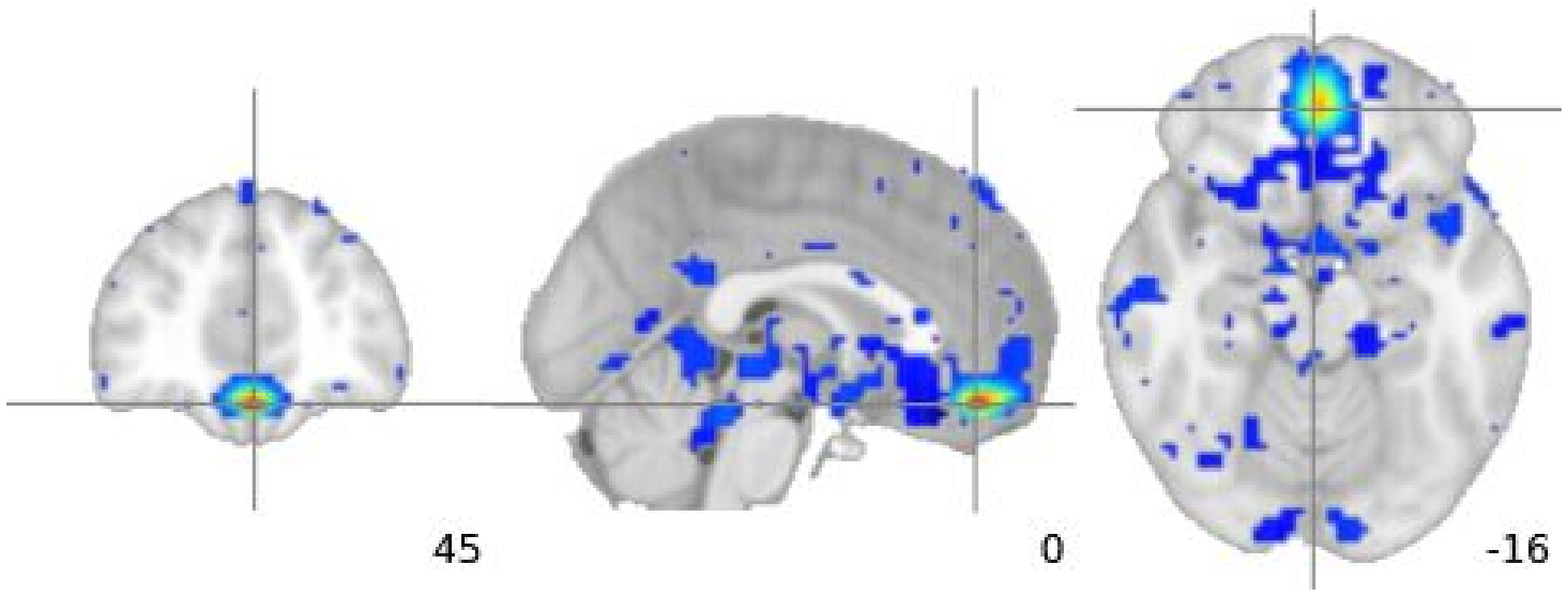} 
    \vskip-1.4em
    {~~\footnotesize\sf\bf (c)}
    &
    \hspace*{-.5ex}\includegraphics[width=\linewidth]{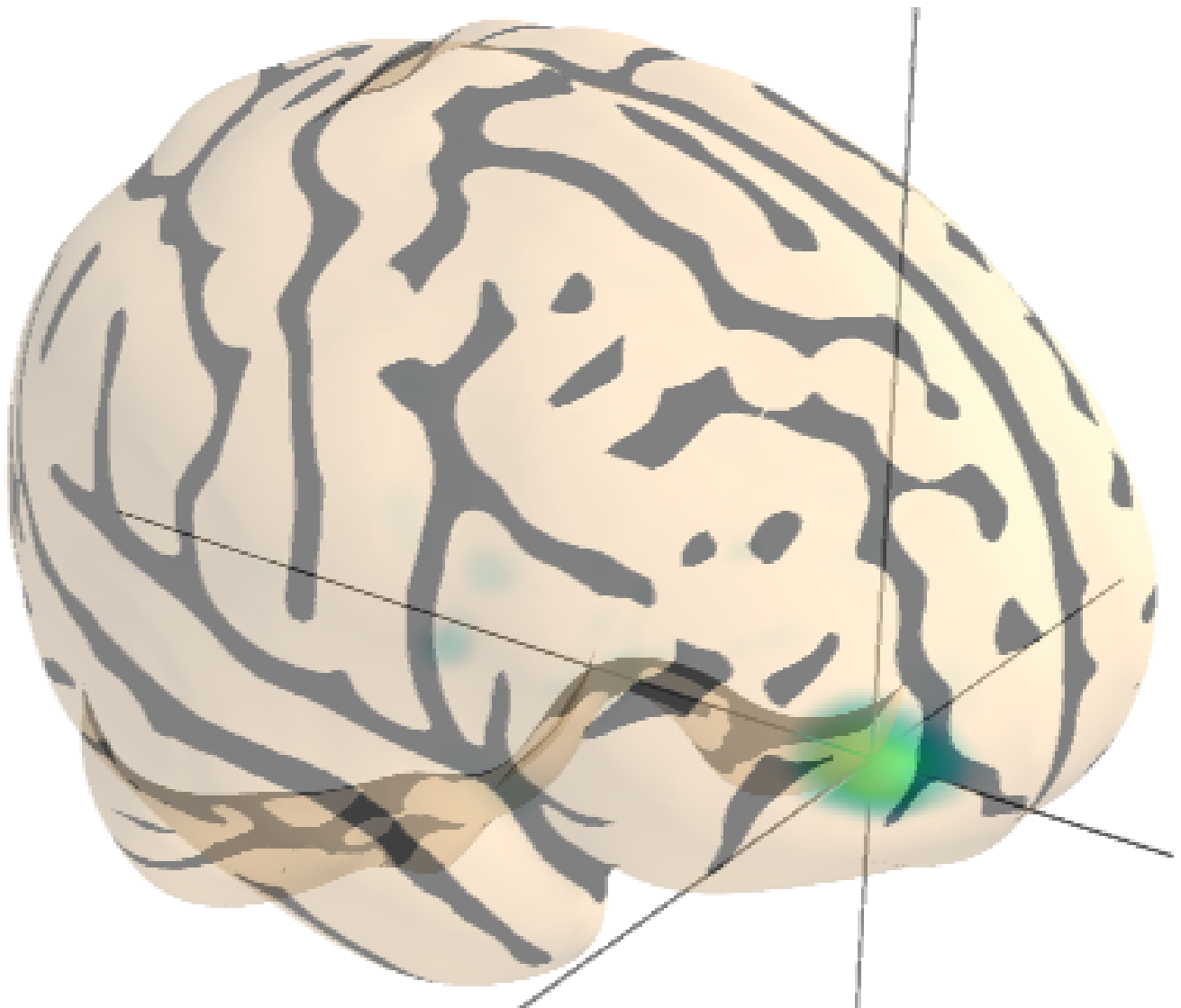} 
    \end{tabular}
  \end{minipage}%
  \hfill%
  \begin{minipage}{.49\linewidth}
    \begin{tabular}{p{.7\linewidth}p{.3\linewidth}}
    \includegraphics[width=\linewidth]{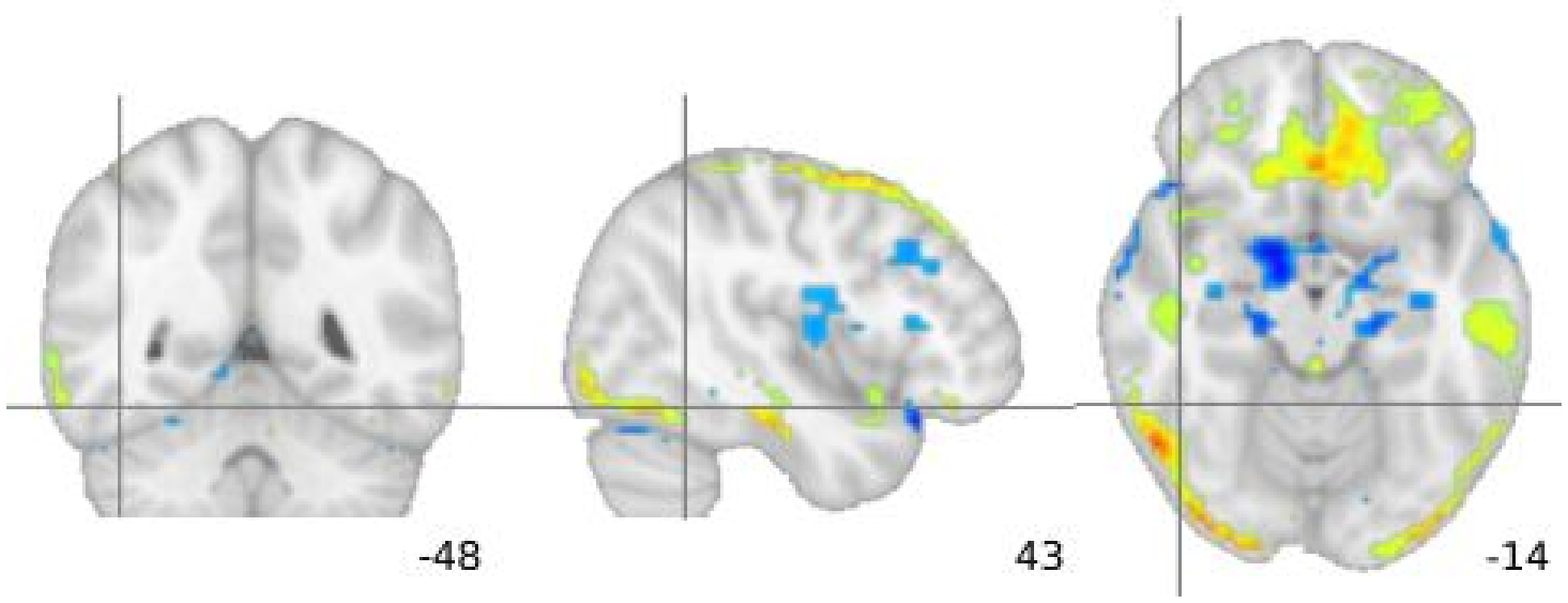}
    \vskip-1.4em
    {~\footnotesize\sf\bf (d)}
    &
    \hspace*{-.5ex}\includegraphics[width=\linewidth]{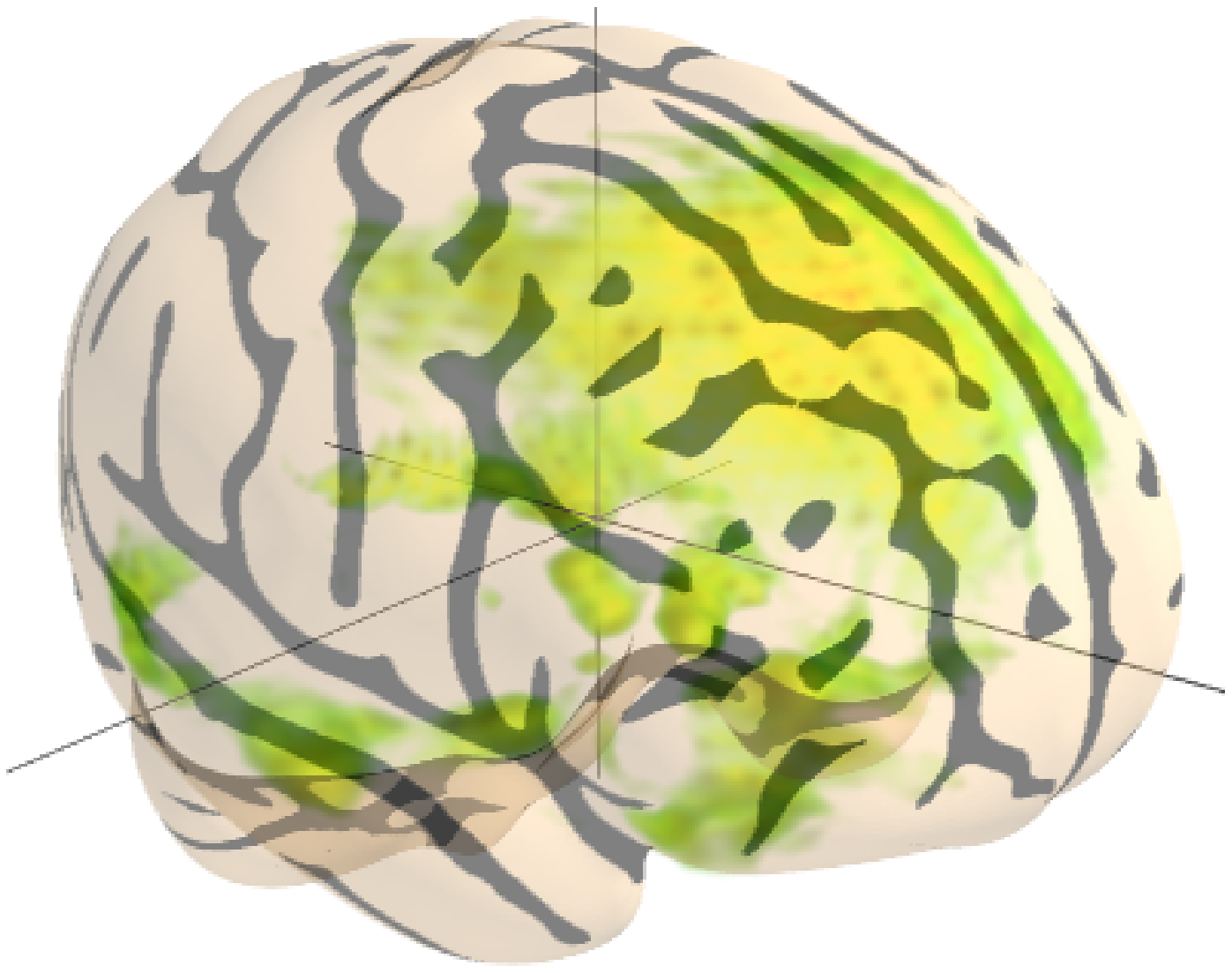}
    \end{tabular}
  \end{minipage}%
  \end{minipage}
   \vskip-2ex
   \caption{
	ICA maps: {\sf\bf (a)} a neuronal component (default mode network),
	          {\sf\bf (b)} a ventricular component,
		  {\sf\bf (c)} and {\sf\bf (d)} physiological noise and
		  motion components.
	\label{fig:ica_maps}
   }

\end{figure}

\begin{figure}[p]
  \begin{center}
    \begin{minipage}{0.8\linewidth}
    \begin{tabular}{p{.65\linewidth}p{.3\linewidth}}
    \includegraphics[width=\linewidth]{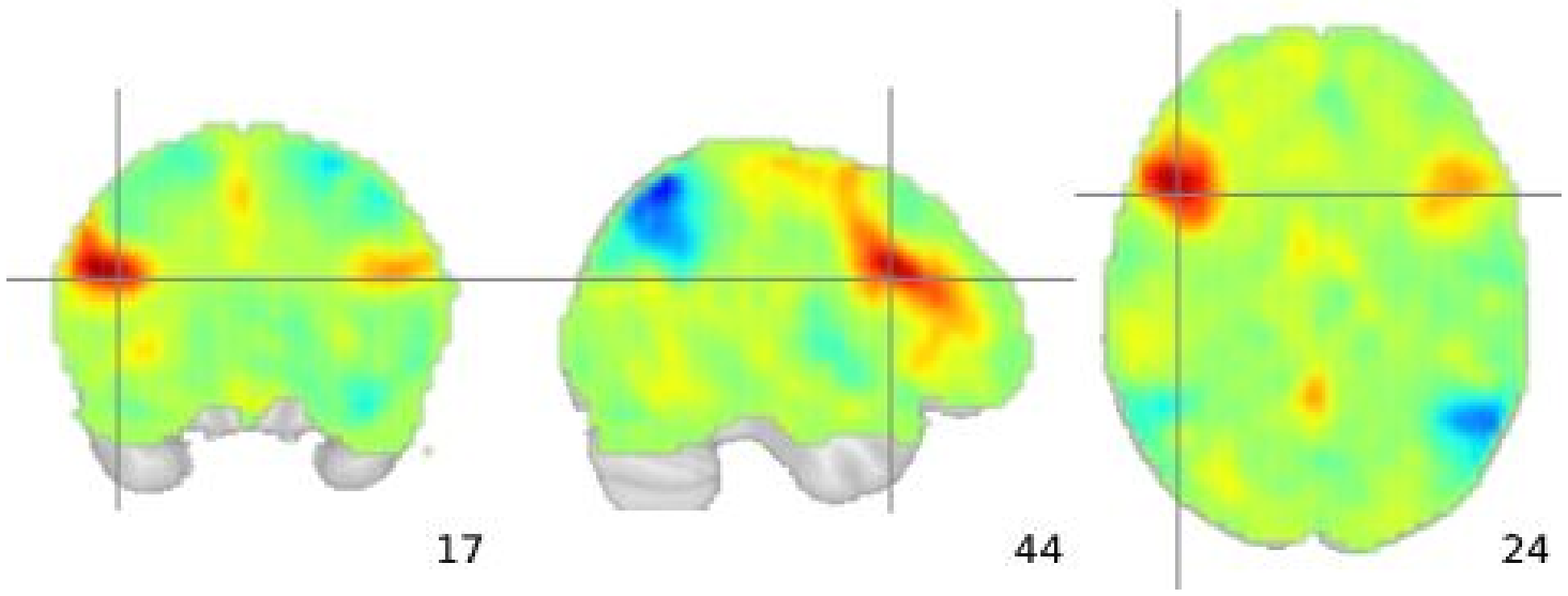}
    \vskip-1.4em
    {\small\sf\bf (a)}
    &
    \includegraphics[width=\linewidth]{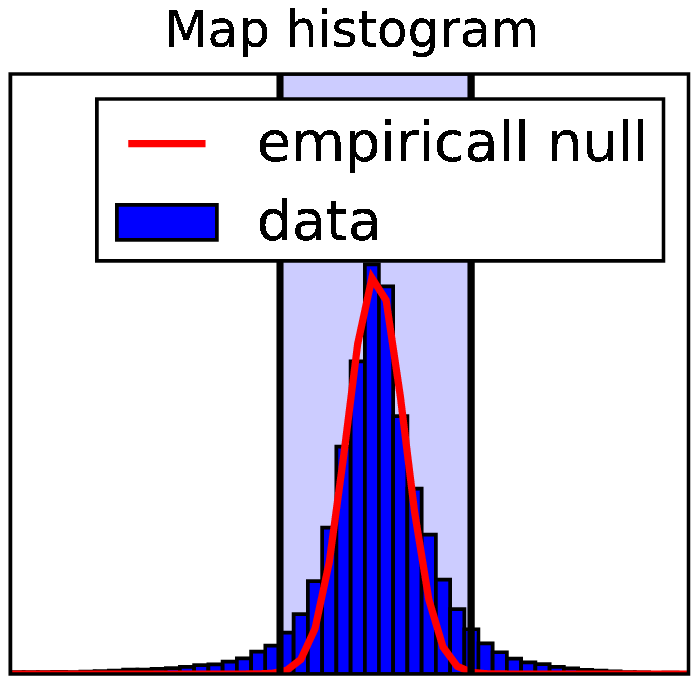}
    \vskip-1.4em
    {\small\sf\bf (b)}
    \end{tabular}
   \vskip-1.2ex

    \begin{tabular}{p{.65\linewidth}p{.3\linewidth}}
    \includegraphics[width=\linewidth]{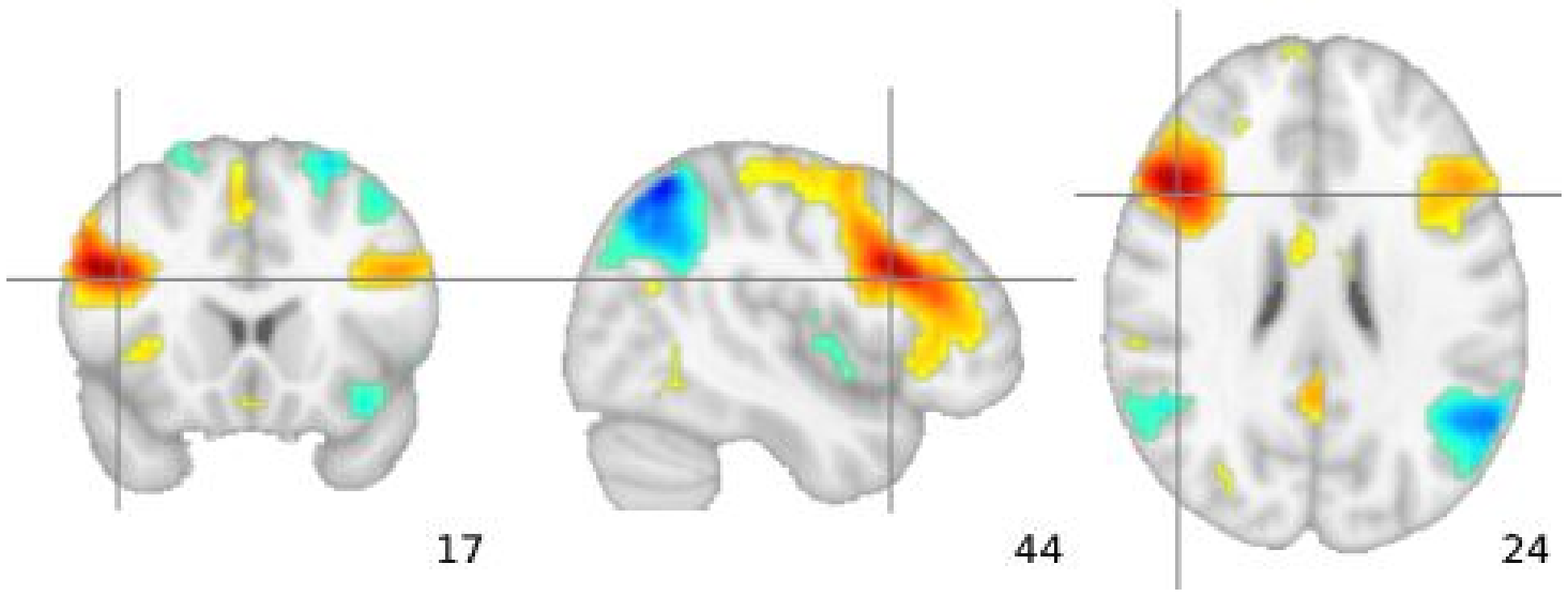} 
    \vskip-1.4em
    {\small\sf\bf (c)}
    &
    \includegraphics[width=\linewidth]{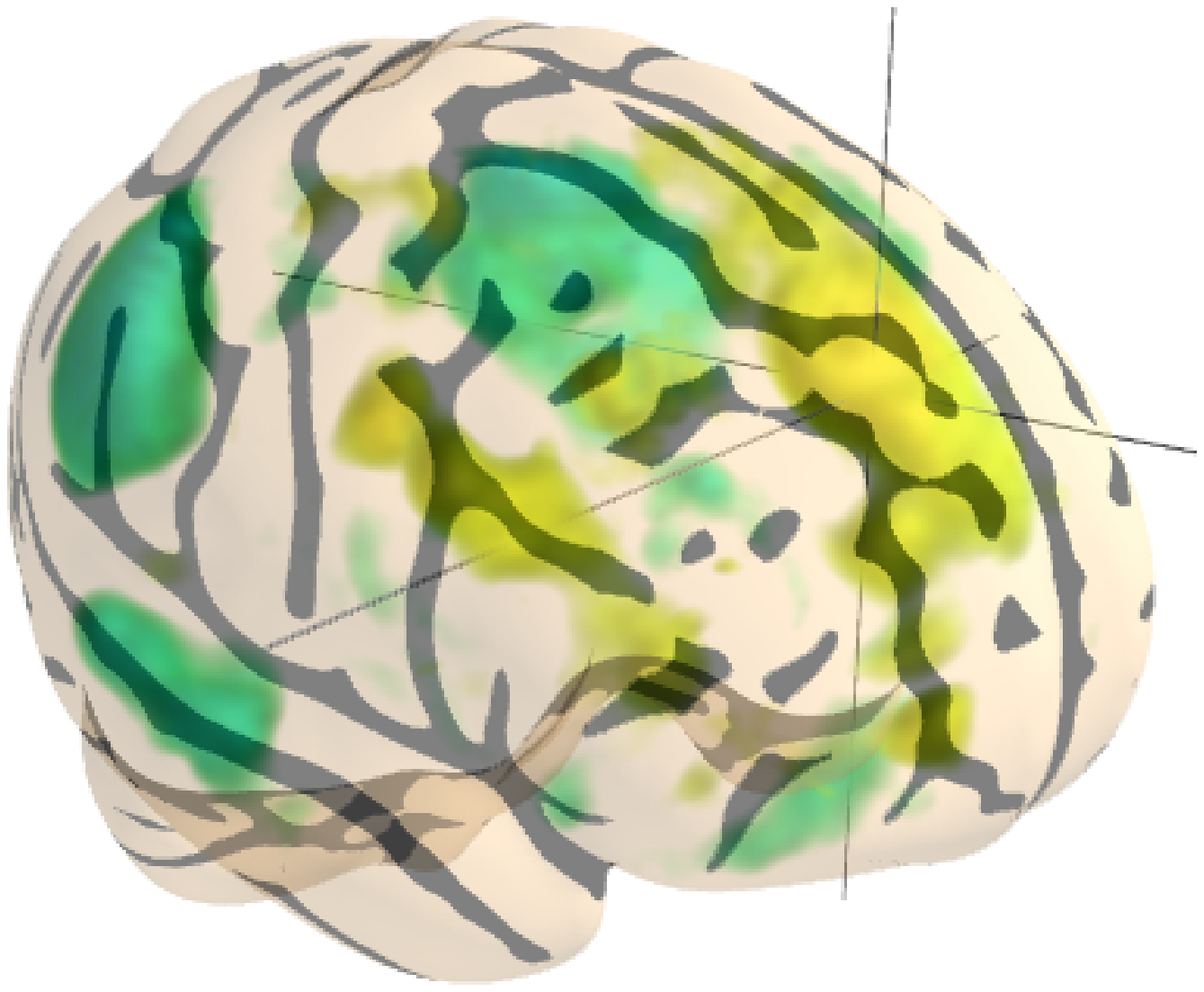}
    \end{tabular}
   \end{minipage}
   \vskip-2.5ex
   \caption{
	{\sf\bf (a)} A neuronal ICA map and {\sf\bf (b)} its histogram. The null
	distribution, shown in red on the histogram, is estimated from the 
	center of the histogram. 
	{\sf\bf (c)} Corresponding thresholded map; only voxels 
	with $p<5{\cdot}10^{-4}$ (uncorrected, two-sided test) are kept.
	\label{fig:thesholding}
   }
  \end{center}
\end{figure}

Twelve healthy volunteers were scanned after giving informed consent. 820
EPI volumes were acquired every $TR = 1.5\,\text{s}$ at a $3\text{mm}$
isotropic resolution, during a rest period of 20 minutes. CanICA
identified 50 non-observation-noise principal components at the subject
level (Eq. \ref{eq:observation_noise}) and a subspace of 42 reproducible
patterns at the group level (Eq. \ref{eq:subject_variability}), which
matches numbers commonly hand-selected by users in current ICA studies.
On these long sequences, model-evidence-based methods such as those used
in \cite{Beckmann2004} select more than 300 components. Extracted maps
can be classified by eye in neuronal components, cardio-spinal-fluid
(CSF) induced fluctuations, and movement-related patterns (Fig.
\ref{fig:ica_maps}). The empirical-null-based thresholding yields best
results for neuronal components (see Fig. \ref{fig:thesholding}). An
interesting side result of these maps is that measurement artifacts such
as movement or CSF noise form reproducible patterns between different
subjects.

\begin{figure}[p]
  \begin{center}
   \centerline{\sffamily\bfseries\small Non-thresholded maps ~}\vskip -0.8ex
   \begin{minipage}{0.6\linewidth}
    \includegraphics[width=\linewidth]{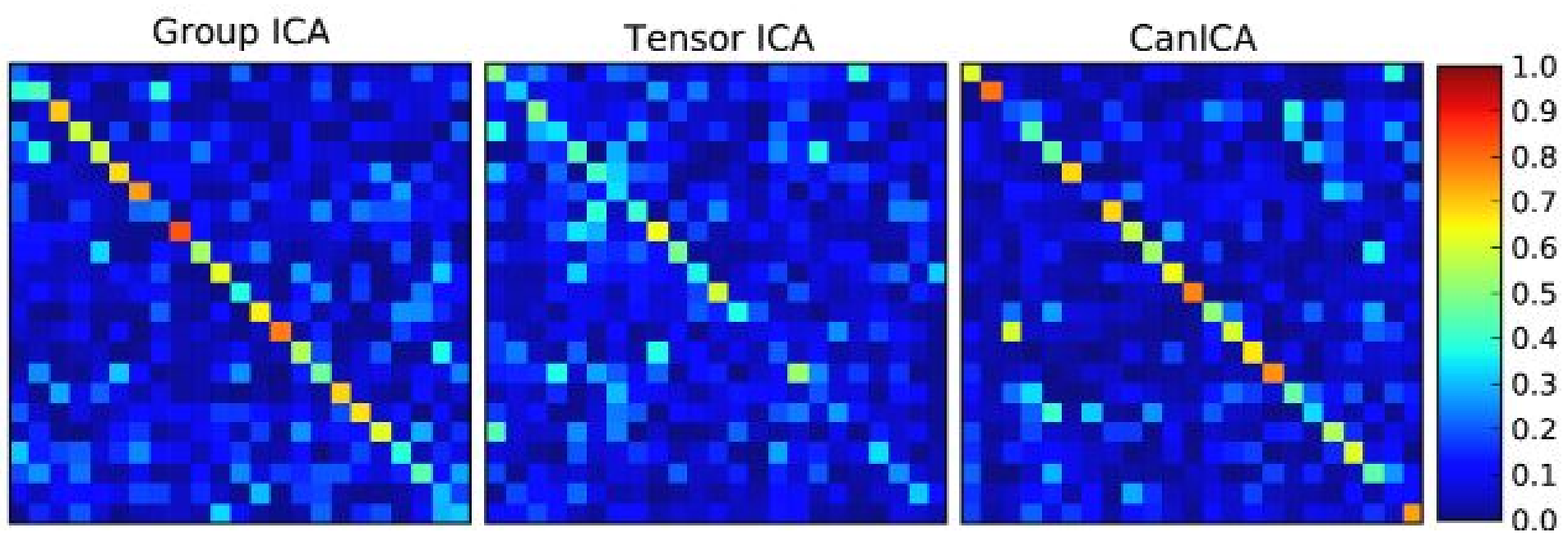}
    \vskip-0.5em
    {\small\sf\bf (a)}
   \end{minipage}%
   \hfill%
   \begin{minipage}{0.38\linewidth}
    \includegraphics[width=\linewidth]{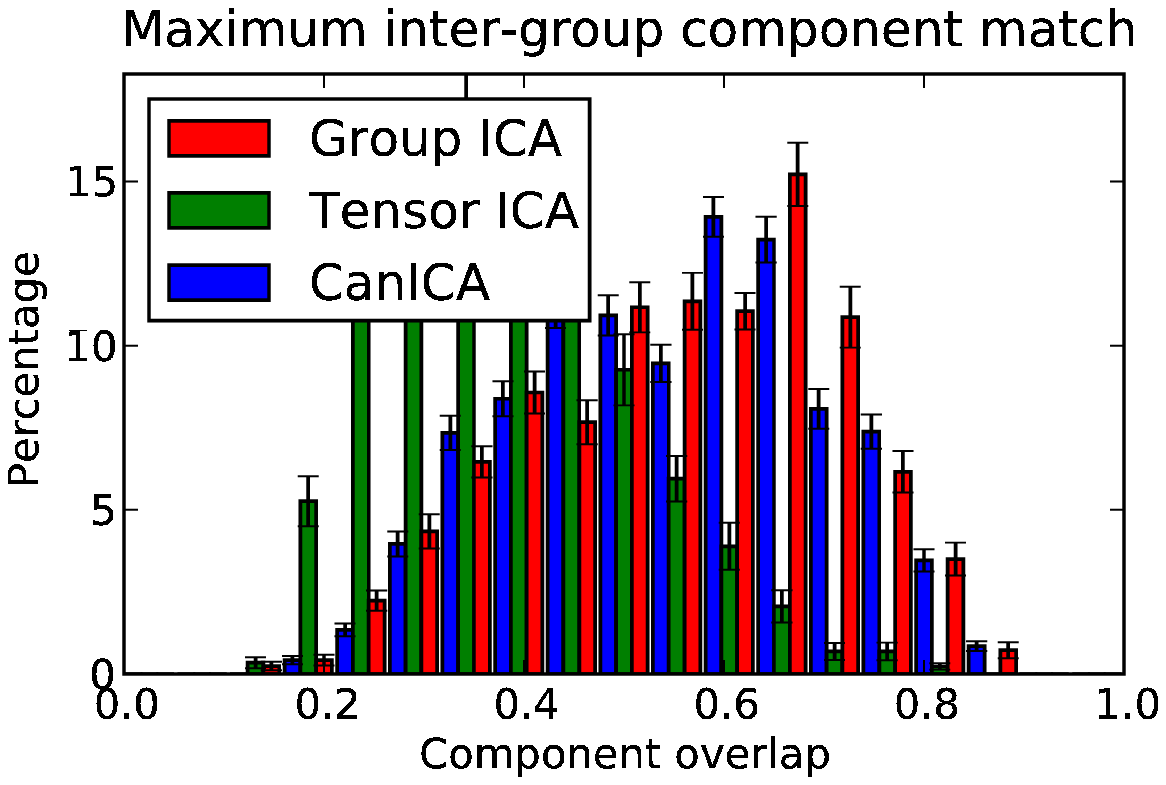}
    \vskip-1.8em\hskip-1ex
    {\small\sf\bf (b)}
   \end{minipage}\vskip-0.2ex
   \centerline{\sffamily\bfseries\small Thresholded maps}\vskip -0.7ex
   \begin{minipage}{0.6\linewidth}
    \includegraphics[width=\linewidth]{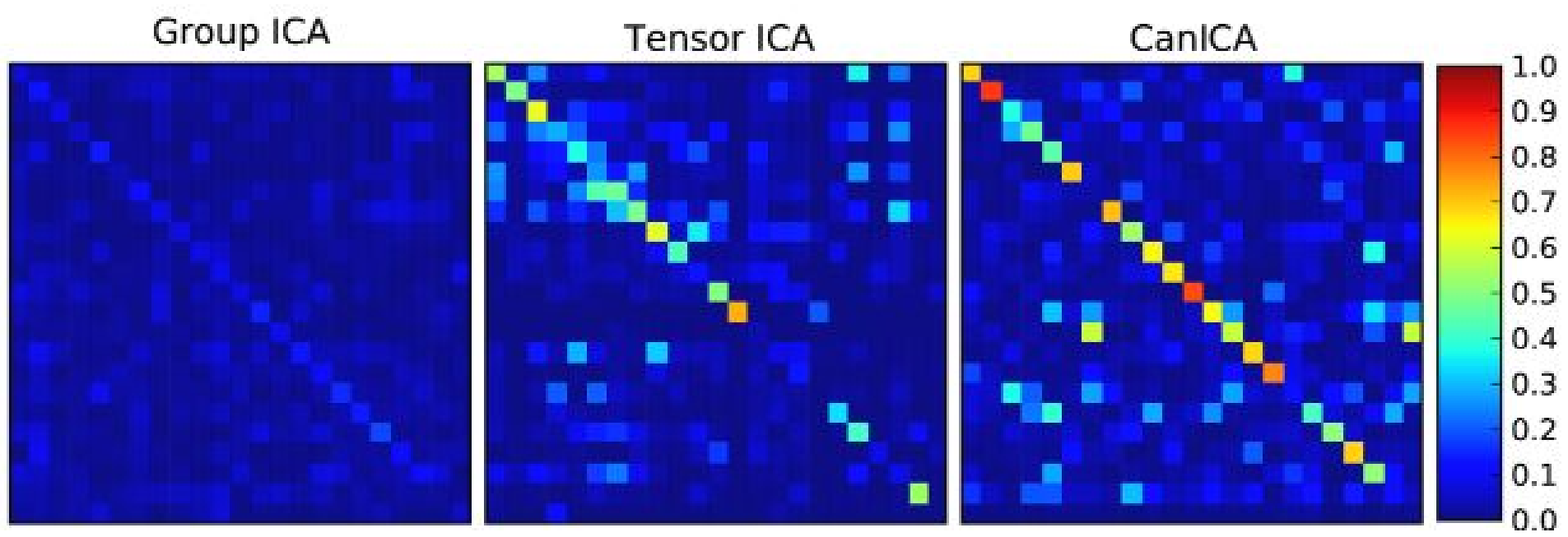}
    \vskip-0.5em
    {\small\sf\bf (c)}
   \end{minipage}%
   \hfill%
   \begin{minipage}{0.38\linewidth}
    \includegraphics[width=\linewidth]{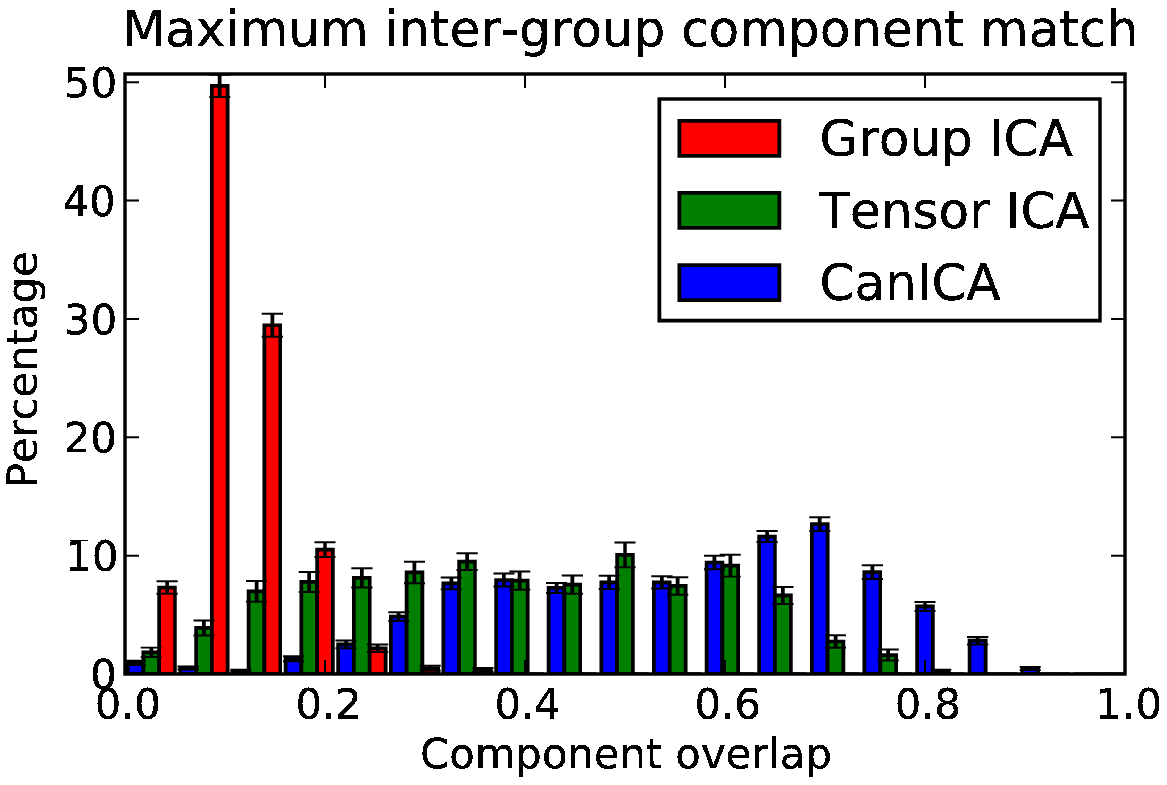}
    \vskip-1.8em\hskip-1ex
    {\small\sf\bf (d)}
   \end{minipage}\vskip-1.1ex
    \caption{
    {\sf\bf (a)}
	Typical cross-correlation matrices between non-thresholded ICA maps 
	learned on two half-splits of the total 12 subjects, for group ICA, 
	tensor ICA, and CanICA.
    {\sf\bf (b)}
	Histogram of the maximal overlap for each component from one
	half-split to a component from the other half.
	\label{fig:split_half_comparison}
    {\sf\bf (c)} and {\sf\bf (d)} cross-correlation matrices 
    and histogram for thresholded ICA maps.
    }
  \end{center}
\end{figure}

\begin{table}[p]
   \begin{center}
    \begin{tabular}{c|c|c|c||c|c|c}
    & \multicolumn{3}{|c||}{Non-thresholded maps} &
    \multicolumn{3}{c}{Thresholded maps} \\
	    & ~ Group ICA ~ & ~ Tensor ICA ~ & ~ CanICA ~
			    & ~ Group ICA ~ & ~ Tensor ICA ~ & ~ CanICA ~ \\
    \hline
    \hline
    $e$	    & 0.58 (0.04) & 0.47 (0.06) & 0.55 (0.05) 
	    & 0.03 (.004) & 0.31 (0.03) & 0.52 (0.05)\\
    $t$	    & 0.53 (0.04) & 0.36 (0.03) & 0.53 (0.05) 
	    & 0.10 (0.01) & 0.35 (0.02) & 0.53 (0.04) 
    \end{tabular}
   \end{center}
   \caption{
	Reproducibility measures $e$ and $t$ for Group ICA, Tensor ICA
	and CanICA calculated on the half-split cross-correlation
	matrices, both for non-thresholded and thresholded maps. Numbers
	in parenthesis give the standard deviation of the mean.
	\label{tab: measures}
   }
\end{table}

\subsection{Inter-group reproducibility}

We performed 38 analyses on paired groups of 6 different subjects. Out of
the 76 groups, 2 yielded 20 stable components, 19 yielded 21 components,
36 yielded 22 components, 17 yielded 23 components, and the last 2
yielded 24 components. We compare our method with tensor ICA
\cite{Beckmann2005a}, running analysis using the MELODIC software on each
group, and group-ICA, using the GIFT ICA toolbox
(\url{http://icatb.sourceforge.net/}). To avoid bias from the
selected-subspace dimension, we run the tensor ICA and group ICA analysis
specifying 23 components. 

We do cross-correlation analysis on the non-thresholded ICA maps, but
also use each implementation's thresholding algorithm to separate the
features of interest. For Group-ICA, the thresholding is done on the $t$
statistics maps generated by the algorithm. As these maps have low
amplitude, thresholding on $|t|>3$ leaves very few selected voxels; we
use $|t|>2$, which yields the same number of selected voxels as those
selected by the two other methods.

On non-thresholded maps, CanICA and Group ICA perform similarly whereas
tensor ICA selects a slightly less stable subspace, and thus yields less
reproducible ICA maps (see Fig. \ref{fig:split_half_comparison} and Tab.
\ref{tab: measures}). Thresholding the maps does not significantly change
the subspace stability ($e$) and map reproducibility ($t$) for CanICA,
but decreases performance for tensor ICA and drastically affects
stability and reproducibility for group ICA.

\section{Discussion}
\label{sec:discussion}

\subsubsection{Importance of capturing inter-subject variability}

The close correspondence between the overlap of the selected subspace
($e$) and independent-component matching quality ($t$) suggests that
identification of the reproducible signal is a key step to identifying
stable independent components. Our method explains only a small fraction
of the total signal variance --less than 50\% for all 12 subjects. This
fraction is selected both on noise-rejection criteria, with
well-specified noise models, and to best take in account inter-subject
variability, by CCA. Indeed, CCA selects linear combinations of subject
patterns that have highest canonical correlations. As the individual
subject components are whitened, each one can contribute no more than 1
to the canonical correlation. Thus high canonical correlations ensure
representation of a large fraction of the group. In addition, the
algorithm minimizes the sum of squares of the total subject-variability
residual, $\B{R}$. The stability of the subspace selected by the GIFT
implementation of group ICA is explained by similar reasons.

We believe that subject variability is not accounted for as well by
tensor ICA because the variability noise is estimated during the
tensorial ICA step, for which statistical significance is hard to
establish. The combination of subject-specific components present in the
final estimated independent components is not guarantied to reflect
multiple-subject contributions and in practice the corresponding
subject-loading vectors are often unbalanced across subjects.

\subsubsection{Residual ICA-patterns instability}

%
%

\begin{figure}[b]
  \hspace*{-.05\linewidth}
  \begin{minipage}{1.1\linewidth}%
  \begin{minipage}{.49\linewidth}
    \begin{tabular}{p{.7\linewidth}p{.3\linewidth}}
    \includegraphics[width=\linewidth]{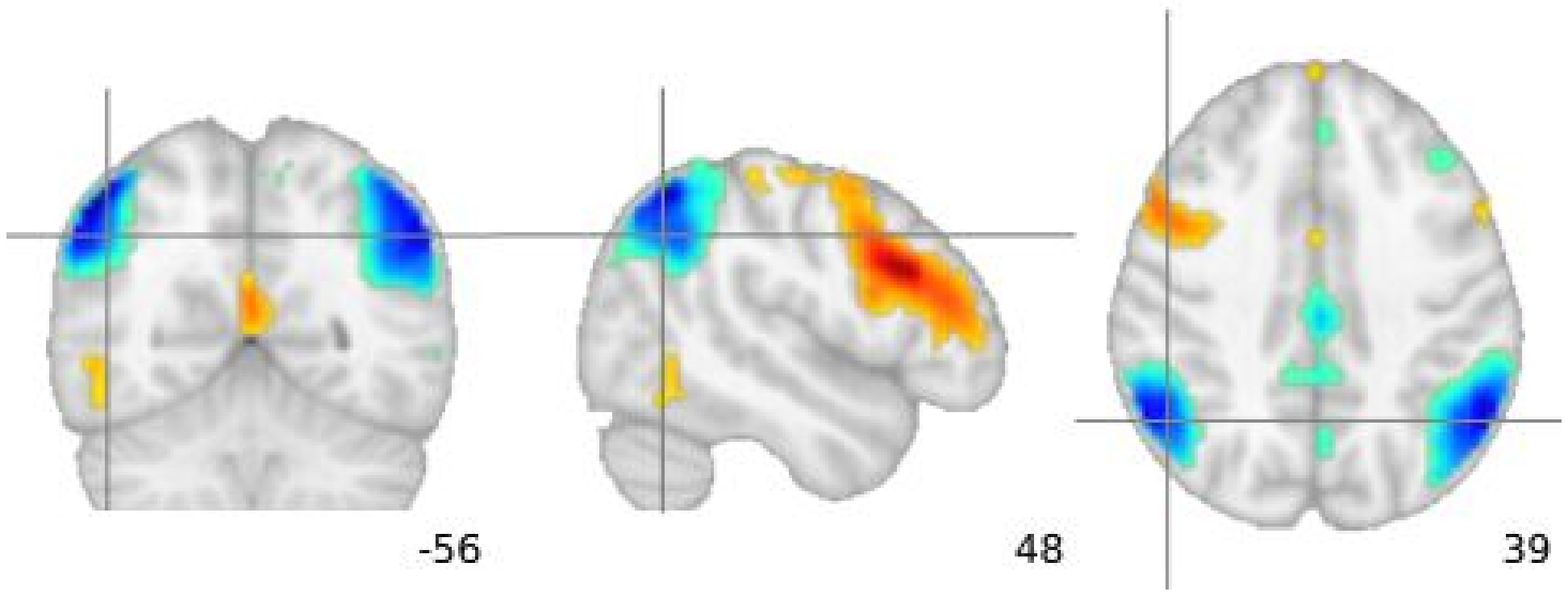} 
    \vskip-1.4em
    {~\footnotesize\sf\bf (a)}
    &
    \hspace*{-.5ex}\includegraphics[width=\linewidth]{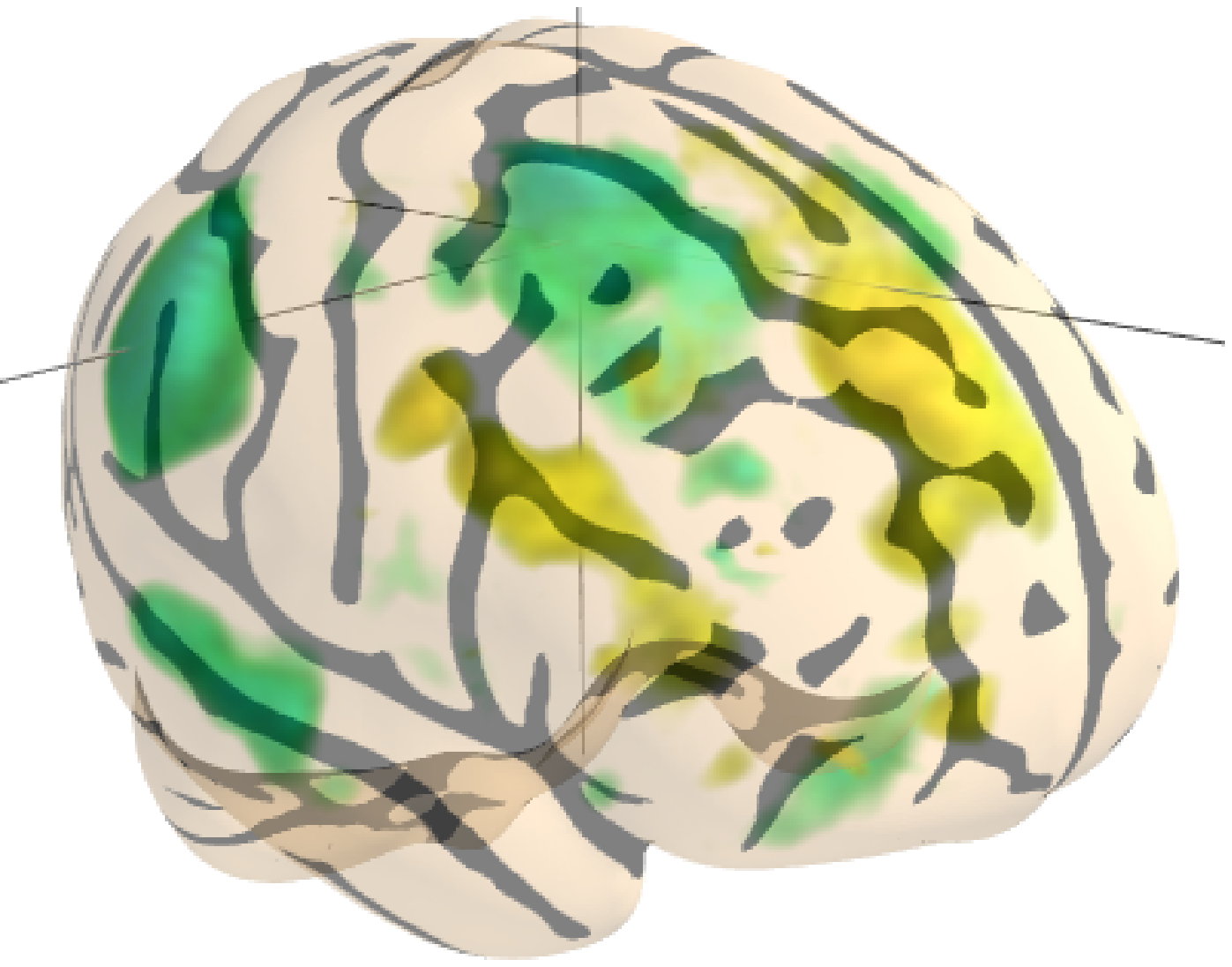} 
    \end{tabular}
  \end{minipage}%
  \hfill%
  \begin{minipage}{.49\linewidth}
    \begin{tabular}{p{.7\linewidth}p{.3\linewidth}}
    \includegraphics[width=\linewidth]{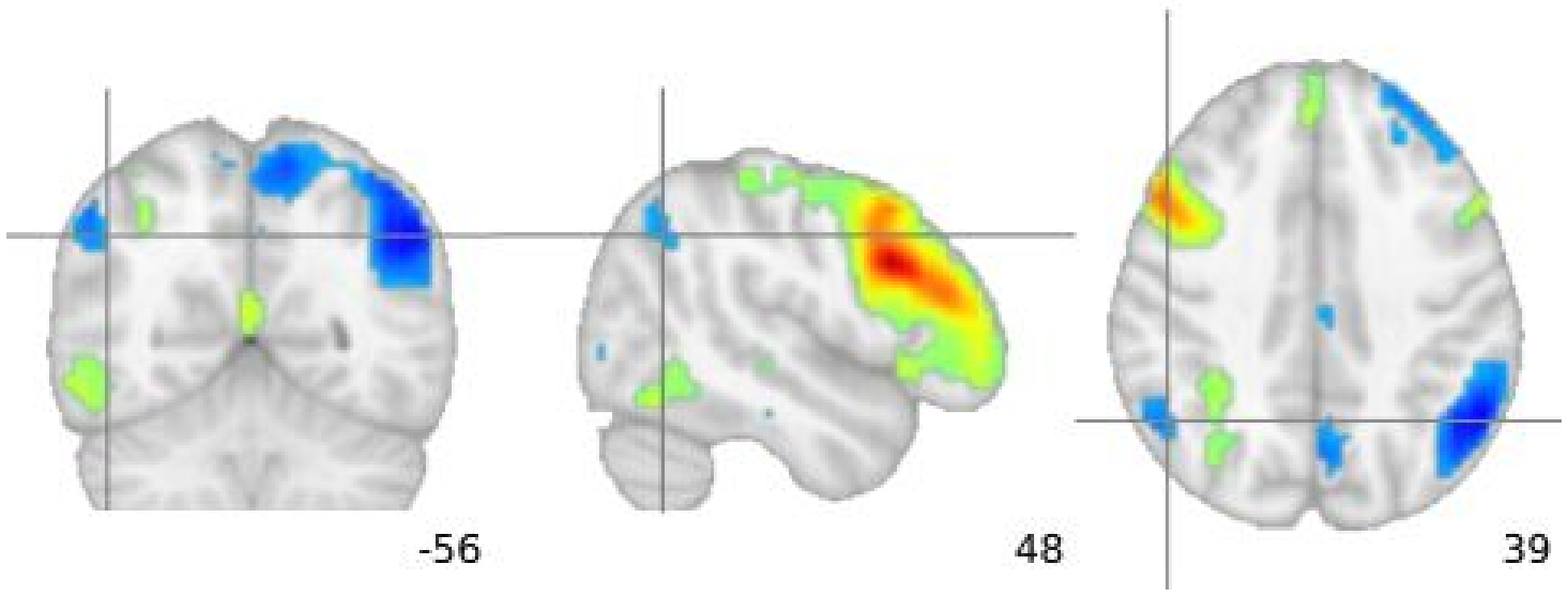} 
    \vskip-1.4em
    {\footnotesize\sf\bf (b)}
    &
    \hspace*{-.5ex}\includegraphics[width=\linewidth]{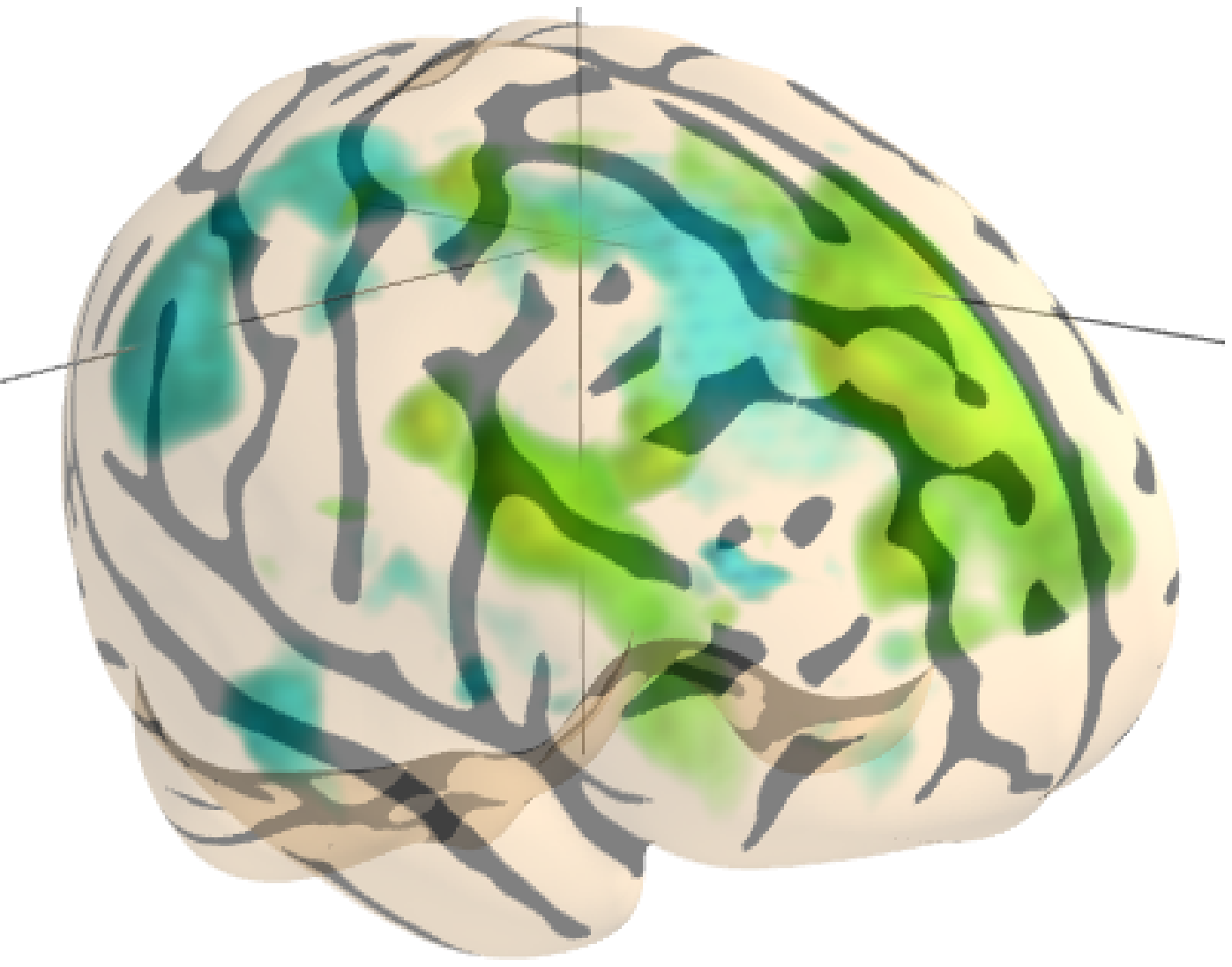} 
    \end{tabular}
  \end{minipage}%
  \end{minipage}%
   \vskip-2ex
   \caption{
	Two ICA patterns corresponding to Fig. \ref{fig:thesholding} estimated 
	from two sub-groups of 11 subjects, with 10 common subjects. On the 
	left, {\sf\bf(a)}, an activation cluster in the left superior 
	parietal lobe is clearly  visible; whereas on the right, 
	{\sf\bf(b)}, the corresponding cluster does not stand out from noise.
	\label{fig:ica_maps_split}
   }
  
\end{figure}

ICA extracts map by successive optimizations of linear combinations to
maximize negentropy. It is very flexible because, unlike PCA, it
does not require orthogonality of the corresponding time courses.
However, even with the careful noise reduction performed by CanICA, small
signal perturbation can lead to different patterns. As an example, CanICA
run on two eleven-subjects groups, differing only by one subject, can
yield patterns in which the plausible neuronal activation clusters differ
(Fig. \ref{fig:ica_maps_split}). One should thus be careful when
inferring cognitive networks using ICA and, when possible, do
significance testing using other criteria, such as seed-voxel correlation
analysis.


%

\section{Conclusion}

We have presented a novel blind pattern-extraction algorithm for fMRI
data. Our method, CanICA, is auto-calibrated and extracts the significant
patterns according to a noise model. From these patterns, reproducible
and meaningful features could be extracted. An important aspect of our
method, specifically designed to perform group analysis, is that the
features selected are more reproducible than other group-level ICA
methods because it identifies a significantly-reproducible signal
subspace and extracts localized features with a criteria consistent with
the ICA algorithm. 

CanICA is numerically efficient, as it relies solely on well-optimized
linear algebra routines and performs the ICA optimization loop on a small
number of components. Performance is important to scale to long fMRI time
series, high-resolution data, or large groups. In addition, as the
group-level pattern extraction (CCA and ICA) is very fast\footnote{a few
minutes for our data on a $2\,\text{GHz}$ Intel core Duo},
cross-validation is feasible. ICA is an unstable algorithm with no
intrinsic significance testing, but we have shown that cross-validation
can be used to establish validity of group-level maps.

\bibliographystyle{splncs}
\bibliography{restingstate}

\end{document}